\documentclass[runningheads]{llncs}
\usepackage{graphicx}
\usepackage{amsmath}
\usepackage{booktabs}

\begin{document}
\title{Medicinal Boxes Recognition on a Deep Transfer Learning Augmented Reality Mobile Application}
\titlerunning{AR Medicinal Boxes Classification}

\author{
    Danilo Avola\inst{1} \and
    Luigi Cinque\inst{1} \and
    Alessio Fagioli\inst{1} \and
    Gian Luca Foresti\inst{2} \and\\
    Marco Raoul Marini\inst{1} \and
    Alessio Mecca\inst{1} \and
    Daniele Pannone\inst{1}
}

\authorrunning{D. Avola et al.}

\institute{
    Sapienza University\\
    \email{\{avola,cinque,fagioli,marini,mecca,pannone\}@di.uniroma1.it}  \and
    Udine University\\
    \email{foresti@di.uniroma1.it}
}

\maketitle             
\begin{abstract} 
Taking medicines is a fundamental aspect to cure illnesses. However, studies have shown that it can be hard for patients to remember the correct posology. More aggravating, a wrong dosage generally causes the disease to worsen. Although, all relevant instructions for a medicine are summarized in the corresponding patient information leaflet, the latter is generally difficult to navigate and understand. To address this problem and help patients with their medication, in this paper we introduce an augmented reality mobile application that can present to the user important details on the framed medicine. In particular, the app implements an inference engine based on a deep neural network, i.e., a densenet, fine-tuned to recognize a medicinal from its package. Subsequently, relevant information, such as posology or a simplified leaflet, is overlaid on the camera feed to help a patient when taking a medicine. Extensive experiments to select the best hyperparameters were performed on a dataset specifically collected to address this task; ultimately obtaining up to 91.30\% accuracy as well as real-time capabilities. 

\end{abstract}

\section{Introduction}
\label{sec:introduction}

In the last decade, machine learning and deep learning algorithms have become widespread tools to address many computer vision-based problems, including medical imaging analysis \cite{avola2021ultrasound}, person re-identification \cite{avola2020bodyprint,avola2022person}, environment monitoring \cite{avola2017real,avola2016multipurpose,avola2021low}, emotion recognition \cite{avola2020deep}, handwriting validation \cite{avola2021r,avola2007ambiguities}, background modeling \cite{avola2019new}, and video synthesis \cite{avola2022human}.
Although effective, these methods usually require a large amount of training data which, however, is not always available. To solve this issue, transfer learning approaches are being exploited to retain previous knowledge from different tasks, and reduce the amount of data required to address the new one \cite{zhuang2020comprehensive,avola2019master}. 
A paradigm that is proving particularly relevant to the medical field, where sensitive data is generally employed. As a matter of fact, automatic procedures for detection, classification, and analysis are already being developed \cite{avola2021multimodal,petracca2015virtual}. What is more, in concert with these diagnosis-related procedures, mobile applications are also being explored to support patients in their daily routine \cite{petersen2017development,konig2018use,tun2021internet} since, in general, dealing with the medical aspects of life is a challenging task for both medical operators and sick persons. In particular, operators have to manage several, often elder, people with different pathologies throughout the day. To help this category, applications noticing anomalous patterns can provide a huge healthcare boost, as shown in \cite{zhao2018real}, where bed falls are detected even at nighttime to safeguard patients. Regarding the latter, instead, they tend to either forget when to take a medicine or its required dosage, as reported by \cite{mayo2018examining}. Although it is possible to look up these instructions on the internet or the patient information leaflet (PIL), it can be difficult to find the correct information; suggesting that there is a technological gap that can be filled.

On a different yet related note, hardware advances are enabling mobile devices to execute ever increasingly complex deep learning algorithms, leading to many performance improvements in different fields such as mobile biometrics security through face \cite{freire2019deep,rios2020deep} and fingerprint \cite{baldini2017survey,rahmawati2017digital} recognition. Moreover, the improved hardware, in conjunction with optimized libraries, allows augmented reality (AR) techniques to be run smoothly on most mobile phones. In general, AR applications add a semitransparent layer on top of the camera feed to provide the user with more information with respect to what is being framed by the device. 
This augmented content can then be decided by analyzing the scene, by exploiting markers, or by using GPS coordinates \cite{de2020survey}. What is more, this technology can be applied in heterogeneous application fields such as rehabilitation \cite{avola2019interactive,avola2013design}, art \cite{he2018art,aitamurto2018impact}, or teaching \cite{turkan2017mobile}, and it is used to make the application more engaging and compelling to the user. 

In this paper, to both leverage hardware advances and help people to take a correct medicinal dosage, we present a mobile AR application that recognizes different kinds of medicines from their package, and provides the user with useful information such as its posology or simplified leaflet. In particular, starting from the smartphone camera feed, video frames are sent to an inference engine for classification. Specifically, the engine is based on a deep neural network exploiting the transfer learning paradigm to overcome the limited amount of training data. Finally, AR is employed to present information on the recognized medicine, and enable real-time interactions by the users. To ensure this time constraint, extensive experiments were performed on a dataset specifically collected for this task, so that the best performing model could be used inside the application.

The rest of this paper is organized as follows. Section~\ref{sec:method} introduces the AR mobile application and inference engine used to classify medicinal boxes. Section~\ref{sec:experiments} summarizes information on the collected dataset, relevant implementation details, as well as the obtained results. Finally, Section~\ref{sec:conclusions} draws some conclusions on the presented work.

\section{Background and Method}
\label{sec:method}

In this section, we present the proposed AR pipeline used to classify medicinal packages, and provide details on the application interface as well as the underlying inference engine exploiting the transfer learning paradigm.

\subsection{Mobile App AR Interface}

\begin{figure}
	\centering
	\includegraphics[width=.7\columnwidth]{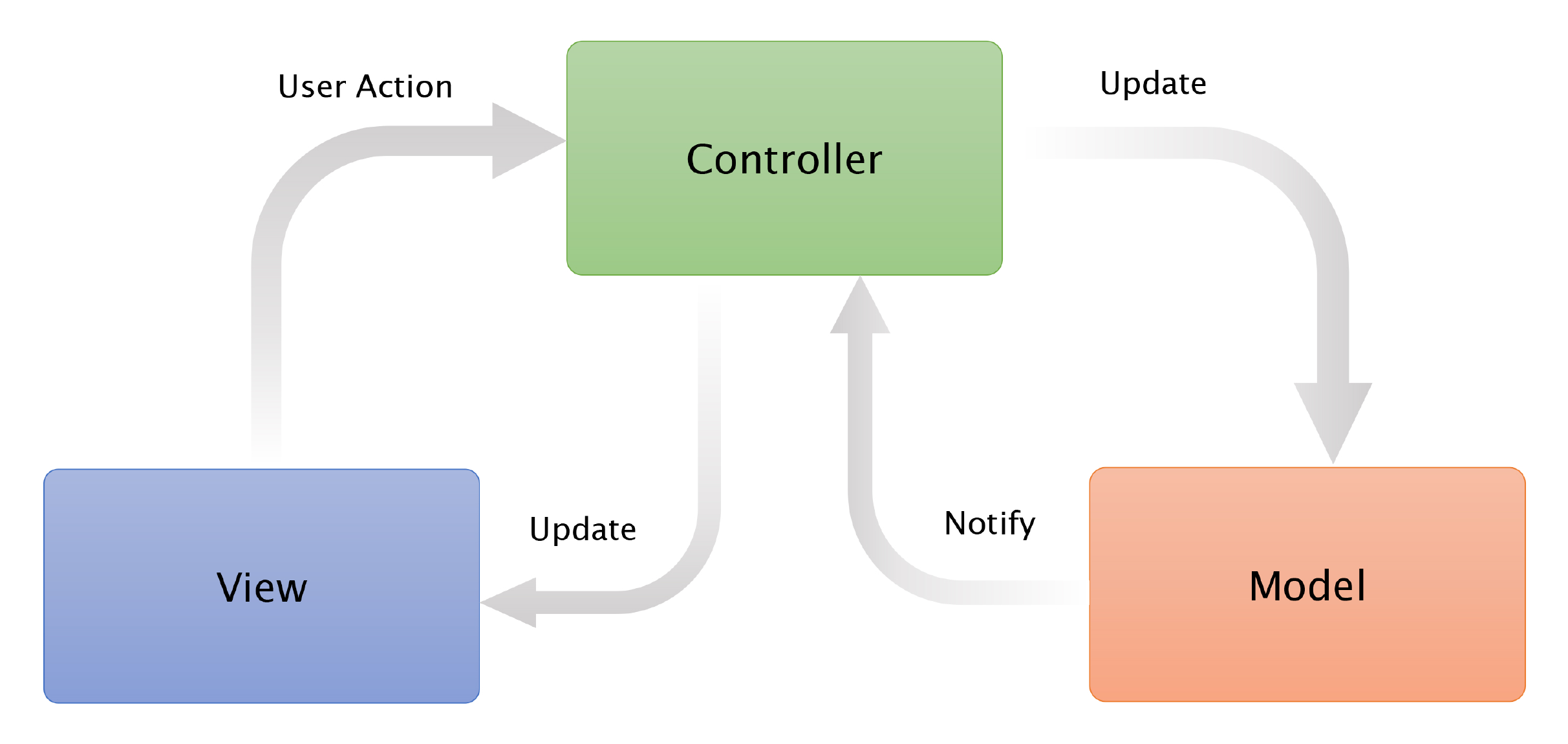}
	\caption{MVC pattern design scheme.}
	\label{fig:mcv_pattern}
\end{figure}

\begin{figure}
	\centering
	\includegraphics[width=0.5\columnwidth]{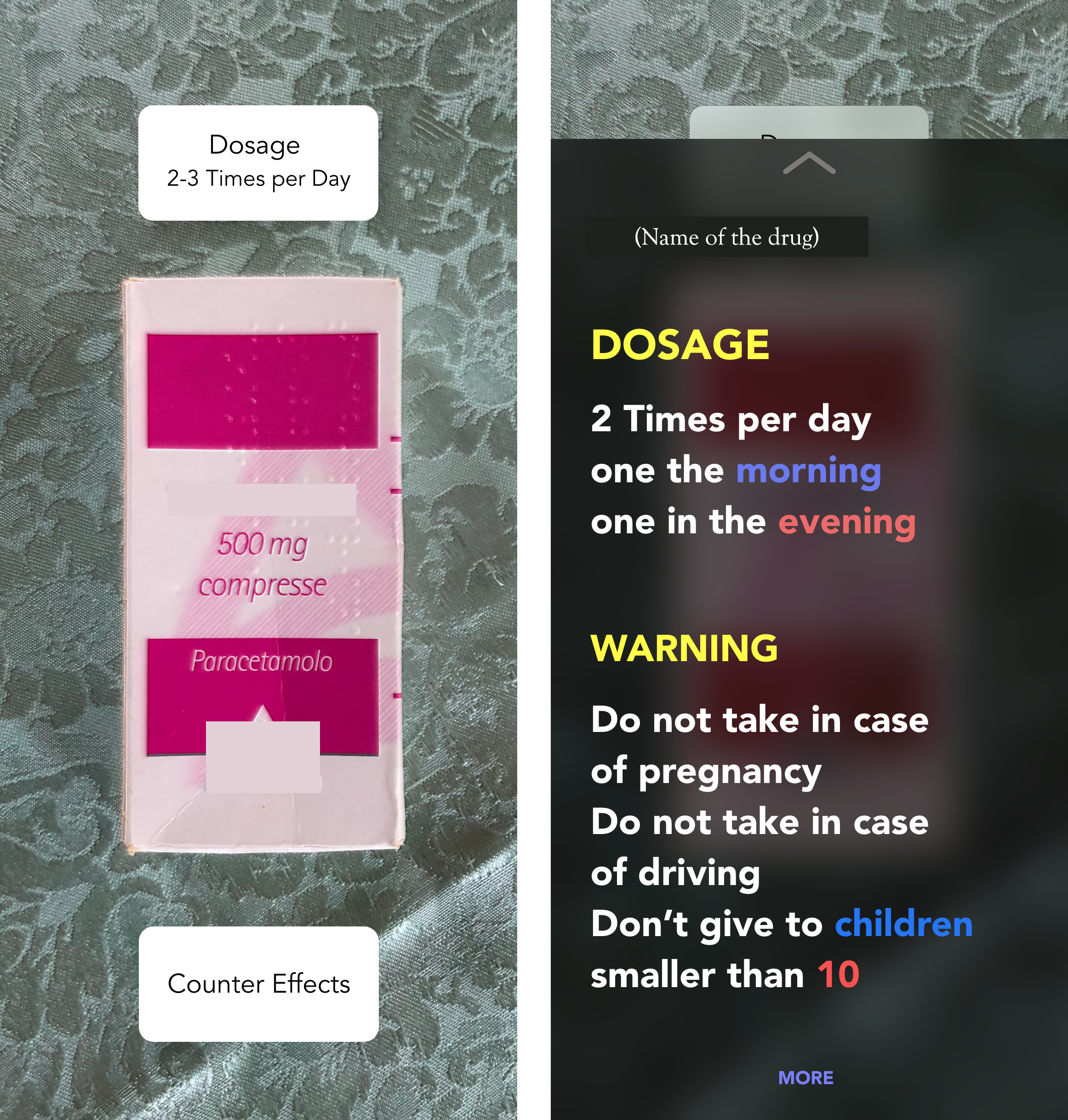}
	\caption{In-app AR user interface screenshots. On the left, the base overlay showing quick details on the recognized medicine posology. On the right, additional PIL extracted information are instead displayed.}
	\label{fig:ui}
\end{figure}

To provide patients with extra information on a given medicine in real-time, we devised an iOS AR application that rapidly shows the suggested posology as well as more detailed PIL guidelines when prompted. The mobile application was implemented following the common model–view–controller (MVC) design pattern, summarized in Fig.~\ref{fig:mcv_pattern}, where different software components are organized into one of the MVC roles. 
Specifically, application logic and operations, such as the inference engine, are associated to the model; interface aspects, including rendered AR elements properties, are linked to the view; while the controller is responsible for the data flow between model and view components, allowing for a responsive and well-organized application.
By following this pattern, it is possible to develop highly re-usable software and it is easier to extend pre-existing libraries. In particular, we exploited the Apple ARKit framework to implement AR capabilities. 
In detail, the application controller component sends video frames captured through the mobile device camera to the inference engine and, whenever a medicinal package is recognized, the view component is updated in real-time so that the RGB camera feed is overlaid with meaningful 2D text information, e.g., posology or PIL. Furthermore, the latter are correctly bounded to the corresponding object through the aforementioned framework functionalities; effectively presenting relevant data of a particular medicine to the user. The intuitive AR interface showing either posology or simplified PIL information is reported in Fig.~\ref{fig:ui}. 

\subsection{Inference Engine}

\begin{figure}
	\centering
	\includegraphics[width=.5\columnwidth]{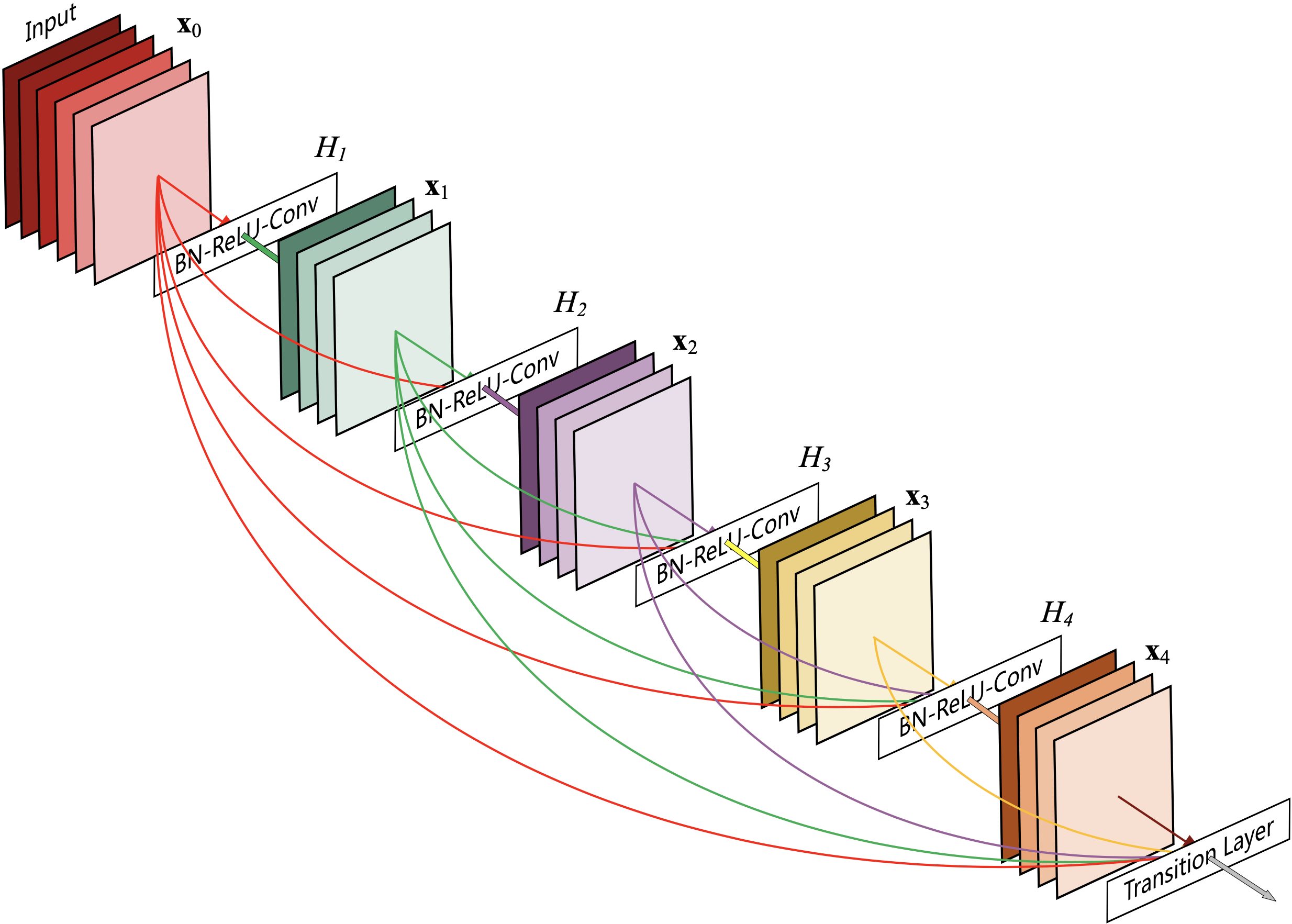}
	\caption{Densenet dense block scheme. Image courtesy of \cite{huang2017densely}.}
	\label{fig:dense_block}
\end{figure}

\begin{figure}
	\centering
	\includegraphics[width=\columnwidth]{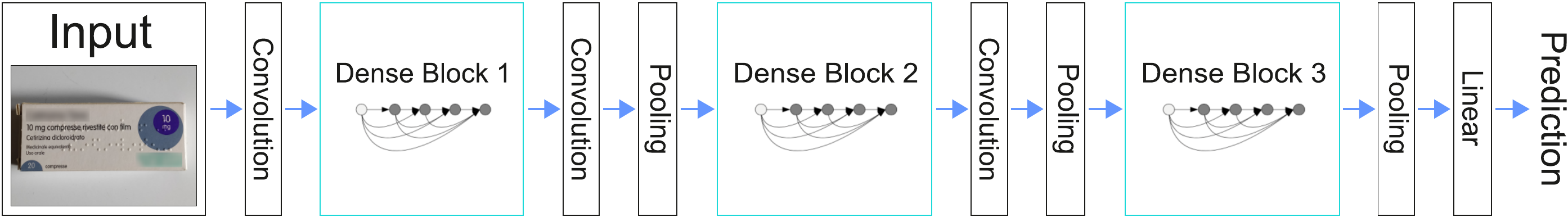}
	\caption{Densenet architecture showing dense blocks interleaved by transition layers.}
	\label{fig:architecture}
\end{figure}

To discriminate between medicinal packages and show the correct information inside the AR interface, the application requires a component to perform inference. The latter was implemented as a deep neural network based on transfer learning and imported as the application inference engine via the iOS CoreML framework. In particular, an ImageNet pretrained densenet \cite{huang2017densely} was selected due to its generally high classification performances as well as its internal architecture. In detail, this network leverages a dense connectivity by introducing links from any layer to any subsequent one, as shown in Fig.~\ref{fig:dense_block}.
Formally, given the $l$-th layer, its input $x_l$ will be composed by the feature maps of all preceding layers, as follows:
\begin{equation}
    x_l = H_l([x_0, x_1, \dots, x_{l-1}]),\label{eq:1}
\end{equation}
where $[x_0, x_1, \dots, x_{l-1}]$ represents the concatenation operation of all previous feature maps; while $H_l(\cdot)$ is a composite function applying batch normalization, a rectified linear unit (ReLU) activation function and a $1\times1$ convolution acting as a bottleneck to consolidate and limit the output size of a given layer. While these dense connections are effective tools to feed-forward information inside a network, they must have the same size in order to be concatenated through Eq.~\eqref{eq:1}. However, this aspect collides with the essential downsampling procedure of a CNN. Therefore, to address this issue, the authors organized their architecture into dense blocks. What is more, transition layers were designed to reduce the feature map size between these blocks. In particular, each transition layer contains a batch normalization operation, a $1\times 1$ convolution, and a $2\times 2$ average pooling layer. Moreover, a compression hyperparameter $\phi$ is also applied on transition layers to increase the architecture compactness by reducing the next block input 
features maps $m$ to $\lfloor\phi m\rfloor$. 
Furthermore, notice that $m$ directly depends on another hyperparameter $k$, called growth rate, that indicates the number of feature maps (i.e., filters) to be produced per layer. Intuitively, if $H_l$ produces $k$ feature maps, the $l$-th layer will have $k_0+k*(l-1)$ maps as input, where $k_0$ corresponds to the input image channels. Thus, by modifying $k$, it is possible to substantially change the number of network parameters. An overview of a densenet with dense blocks and transition layers is shown in Fig.~\ref{fig:architecture}.

To apply the transfer learning paradigm and retain previous knowledge computed on the ImageNet dataset, all model weights are frozen apart from the linear layer used as a classifier component. The latter is then modified by changing its size to handle the right number of classes. Subsequently, the classifier is trained on the medicinal package classification task using the densenet extracted feature maps via the classical backpropagation algorithm and cross-entropy loss, thus exploiting prior knowledge via the frozen layers.  

Finally, since the model receives frames from the camera feed in real-time, a $\lambda$ threshold is applied on the confidence scores produced by the densenet to discard all classifications below it, to avoid showing the user information on uncertain recognitions. Indeed, by using this strategy, possible misclassification of objects with similar shapes, e.g., a generic box, are ignored by the application and only medicinal packages will be enhanced through the proposed AR interface. 

\section{Experiments}
\label{sec:experiments}

In this section, we first introduce the dataset used to test the mobile AR application, which was specifically collected to address the medicinal box classification task. Subsequently, implementation details for all of the described technologies are reported. Finally, a discussion on summarized performances obtained by the inference engine is presented. 

\subsection{Dataset}

\begin{figure}
	\centering
	\includegraphics[width=\columnwidth]{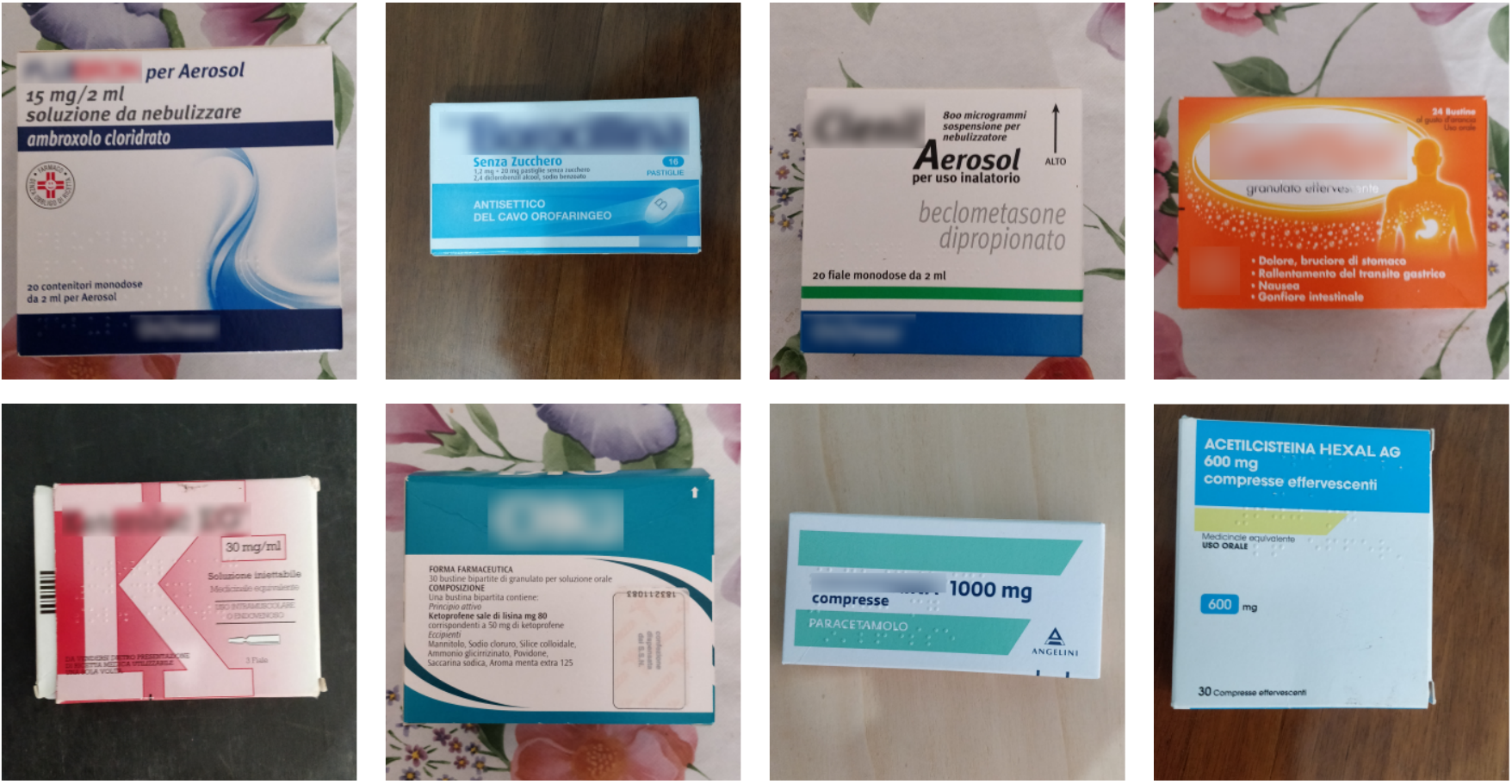}
	\caption{Medicinal boxes samples from the collected dataset.}
	\label{fig:dataset}
\end{figure}

To correctly assess the presented model, 978 images for 63 distinct medicinal packages were collected from different sources such as Google and Bing images, as well as offline, directly through the phone camera. 
In addition, to reinforce the dataset difficulty, we ensured that among the 63 categories there were several distinct boxes presenting various similarities, e.g., shape or color scheme.
Moreover, different lighting conditions, as well as distance and angles from the camera, were captured to further increase data variability. 
To train and test the model, the collection was divided using a stratified 10-fold cross-validation procedure using $80\%/20\%$ splits for the train and test sets, respectively. Notice that this approach retains the requested number of samples per class, thus ensuring there are enough samples for both training and test phases. Moreover, to further improve the model abstraction capabilities, a data augmentation strategy was devised by means of random horizontal flips and random rotations $\theta\in[-15,15]$ degrees. 
Samples from the collected dataset are shown in Fig.~\ref{fig:dataset}.

\subsection{Implementation Details}

The presented mobile application was developed on an iOS mobile phone, i.e., iPhone 12 Pro. This device implements ARKit $5.0$ and CoreML $4.1$ frameworks, which enable, respectively, AR capabilities and neural network translation as well as its execution.
This phone was selected due to both its ability to run the latest frameworks versions, as well as its hardware specifications, which allowed for a smooth real-time AR experience. 

Concerning the transfer learning procedure, before implementing the application inference engine, the densenet was developed using the PyTorch framework and torchvision library, which contains an ImageNet pre-trained densenet version. In particular, we selected the densenet-121 model and modified its last linear layer to have 63 nodes, to match the dataset classes. Moreover, the default growth rate $k=32$ and compression $\phi=0.5$ were set for the architecture. A $\lambda=0.85$ was also found to be a good confidence threshold to avoid uncertain classification showing in the AR interface. 
Regarding the training hyperparameters, the model was fine-tuned on the collected dataset for 100 epochs, using the SGD algorithm, with a learning rate $lr$ set to $0.1$, decreased by a factor of 10 at epochs 40, and 80, a weight decay of 5e-4, and a Nesterov momentum of $0.9$. 

Finally, both training and experiments were performed using a 6-Core Intel i7 2.60GHz CPU with 32GB RAM and single GPU, i.e., a GeForce GTX 1070 with 8GB of dedicated RAM.

\subsection{Performance Evaluation}

In order to choose the best model parameters, ablation studies were performed on hyperparameters $k$ and $\phi$. Common classification metrics, i.e., accuracy, precision, recall, and f1-score, were used to assess each configuration that was tested using the aforementioned 10-fold cross-validation procedure. 

Concerning the growth rate parameter $k$, the obtained results are summarized in Table \ref{tab:k_ablation}. As shown, by increasing the number of filters generated by each convolutional layer, performances improve across all metrics, reaching up to an average $91.35\%$ f1-score. More interestingly, using a low $k$ number does not allow to represent and learn the input data distribution, resulting in consistent $30\%$ gaps. For higher $k$ values, instead, there are diminished performance returns. As a matter of fact, there is roughly a $0.02\%$ difference between models using $k=32$ and $k=64$, indicating that enough information is already captured by the former value. What is more, the number of parameters and memory used at inference time increase superlinearly with respect to the growth rate $k$. 
Thus, since there are negligible performance gains, it is important to choose the correct $k$ value to reduce the mobile device computational burden during inference; resulting in $k=32$ as the better choice for the presented work.
\begin{table}
    \centering
    \begin{tabular}{l|crr|cccc}
        \toprule
        Model & $k$ & Params & Memory & Accuracy & Precision & Recall & F1-score \\
        \midrule
        densenet-121 & 4 & 0.2M & 0.9MB & 61.13\% & 64.40\% & 61.13\% & 61.89\% \\
        densenet-121 & 8 & 0.6M & 2.4MB & 79.90\% & 80.74\% & 79.90\% & 80.09\% \\
        densenet-121 & 16 & 1.9M & 7.7MB & 89.07\% & 89.42\% & 89.07\% & 89.14\% \\
        densenet-121 & 32 & 7.0M & 28.4MB & 91.30\% & 91.51\% & 91.30\% & 91.33\% \\
        densenet-121 & 64 & 27.3M & 109.7MB & 91.32\% & 91.55\% & 91.32\% & 91.35\% \\
        \toprule
    \end{tabular}
    \caption{Ablation study on growth rate hyperparameter $k$. Results correspond to the 10-fold cross-validation scores average. For all networks, $\phi$ was set to 0.5.}
    \label{tab:k_ablation}
\end{table}

Regarding the compression rate $\phi$, results for various values are summarized in Table \ref{tab:phi_ablation}. As can be seen, reducing or increasing the extracted feature maps on transition layers via $\phi$, has relatively little impact on the number of parameters, memory consumption, as well as all metrics. Nevertheless, using the default $0.5$ value allows to achieve the best performances since first, it avoids losing too much information from the preceding dense block with respect to a lower $0.1$ size; and second, it improves the model abstraction capabilities without incurring in possible overfitting scenarios, as opposed to the $\phi=1.0$ case where performances begin to deteriorate. 
\begin{table}
    \centering
    \begin{tabular}{l|crr|cccc}
        \toprule
        Model & $\phi$ & Params & Memory & Accuracy & Precision & Recall & F1-score \\
        \midrule
        densenet-121 & 0.1 & 6.5M & 26.2MB & 90.00\% & 90.28\% & 90.00\% & 90.06\% \\
        densenet-121 & 0.5 & 7.0M & 28.4MB & 91.30\% & 91.51\% & 91.30\% & 91.33\% \\
        densenet-121 & 1.0 & 7.7M & 31.2MB & 91.26\% & 91.51\% & 91.26\% & 91.31\% \\
        \toprule
    \end{tabular}
    \caption{Ablation study on compression hyperparameter $\phi$. Results correspond to the 10-fold cross-validation scores average. For all networks, $k$ was set to 32.}
    \label{tab:phi_ablation}
\end{table}

\begin{figure}
    \centering
    \includegraphics[width=0.45\columnwidth]{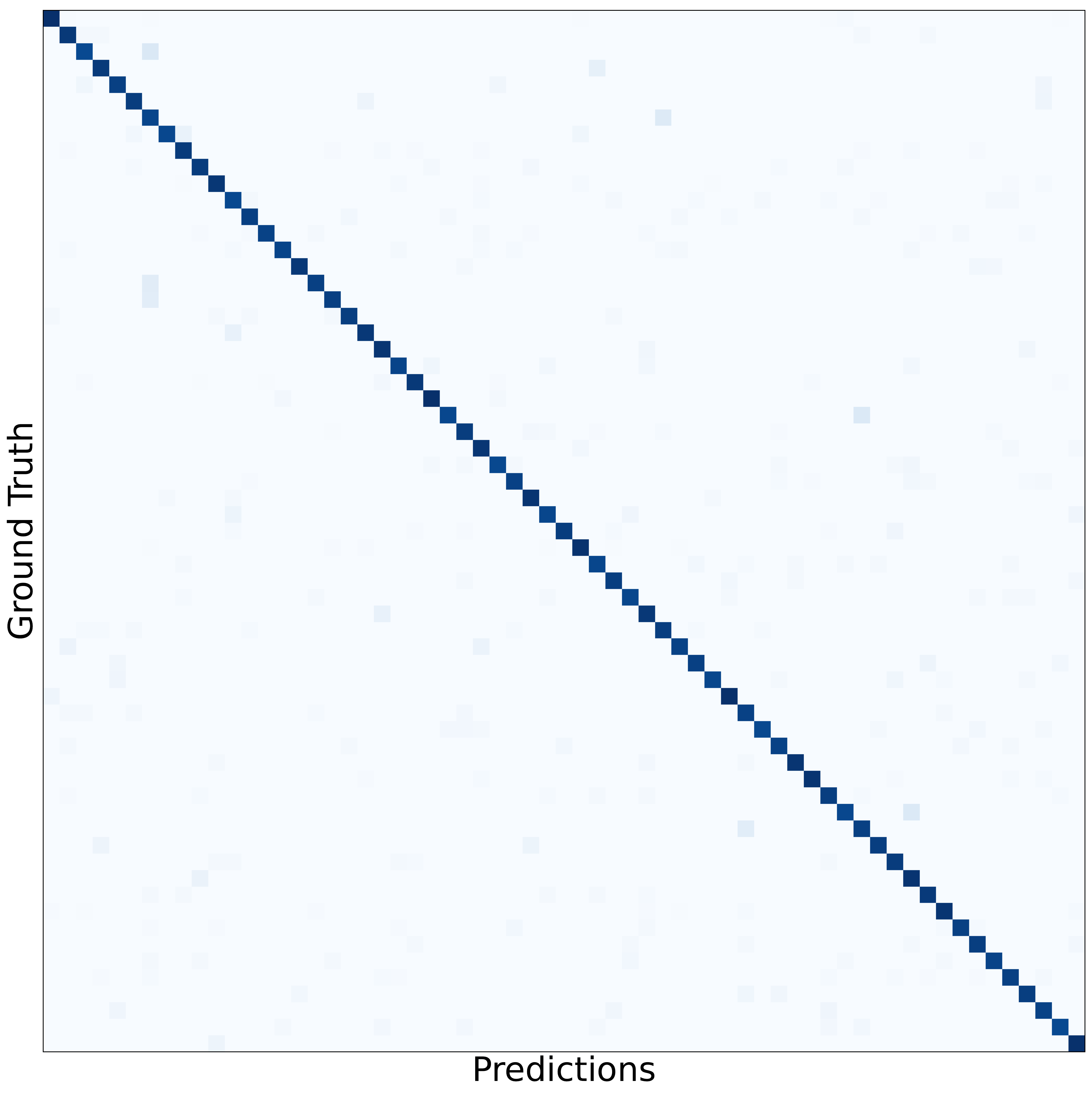}
    \caption{Confusion matrix of a densenet-121 model with $k=32$ and $\phi=0.5$.}
    \label{fig:confusion_matrix}
\end{figure}

\begin{figure}
    \centering
    \includegraphics[width=\columnwidth]{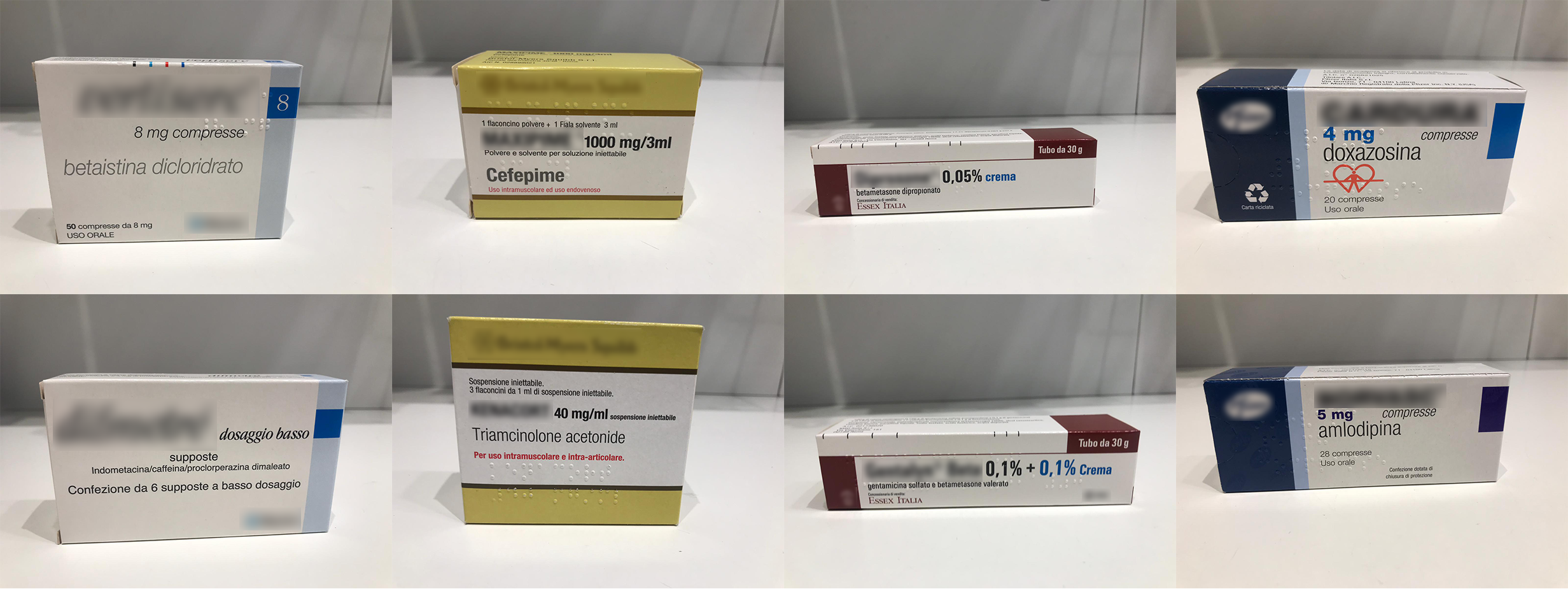}
    \caption{Examples of misclassified medicine boxes.}
    \label{fig:qualitative}
\end{figure}

Although the ablation studies present interesting information, a better system overview can be provided by a confusion matrix, which is reported in Fig.~\ref{fig:confusion_matrix}. 
As expected, most predictions lie in the matrix diagonal, indicating that the model can correctly recognize all classes with a high accuracy. Furthermore, for a given box, misclassifications tend to concentrate on specific medicines, suggesting that there are similarities between those packages.
As a matter of fact, this outcome can be confirmed by observing Fig.~\ref{fig:qualitative}, which shows boxes misclassified by the presented densenet. As can be observed, the reported packages have similar sizes and colors but can have different product names, active substance weights or molecules. This indicates that while the system uses the entire box to recognize a medicine, it is still not able to accurately discriminate text; suggesting that the model can be further improved by refining its ability to classify written content, a task left as future work.

Finally, the densenet model with $k=32$ and $\phi=0.5$ was employed as inference engine inside the mobile application to enable stable real-time capabilities over a 30FPS video stream. Notice that the selected device achieved similar performances even with the more demanding $k=64$ model, however since the AR interface showed several, albeit infrequent, frame drops, the less complex model was still the preferred choice to reduce the device burden.
\section{Conclusions}
\label{sec:conclusions}

In this paper we presented an AR mobile application that can classify medicinal boxes and show the user useful information, such as the posology or PIL, to help them using medicinals in a proper way. Extensive experiments were performed on a dataset specifically collected to address this task, and the best model configuration was selected to achieve real-time execution on the mobile device. Specifically, the chosen architecture, i.e., a densenet, obtains significant performances by exploiting the transfer learning paradigm, reaching up to 91.30\% accuracy on 63 classes; highlighting the effectiveness of the proposed methodology.

\section*{Acknowledgments}
This work was supported by the MIUR under grant “Departments of Excellence 2018–2022” of the Sapienza University Computer Science Department and the ERC Starting Grant no. 802554 (SPECGEO).

\bibliographystyle{splncs04}
\bibliography{mybibliography}
\end{document}